\documentclass[journal]{IEEEtran}

\usepackage{amsmath,amsfonts}
\usepackage{array}
\usepackage[caption=false,font=normalsize,labelfont=sf,textfont=sf]{subfig}
\usepackage{textcomp}
\usepackage{stfloats}
\usepackage{url}
\usepackage{verbatim}
\usepackage{graphicx}
\def\BibTeX{{\rm B\kern-.05em{\sc i\kern-.025em b}\kern-.08em
    T\kern-.1667em\lower.7ex\hbox{E}\kern-.125emX}}
\usepackage{balance}

\usepackage{color}
\usepackage{multirow}
\usepackage[noend]{algpseudocode}
\usepackage[ruled, vlined, linesnumbered]{algorithm2e}%
\usepackage{makecell}
\usepackage{xcolor}
\usepackage{colortbl}
\usepackage{paralist}
\usepackage{amssymb}
\usepackage[pagebackref=true,breaklinks=true,colorlinks,bookmarks=false]{hyperref}
\graphicspath{ {./Figs/} }

\begin{document}

\title{CrossDF: Improving Cross-Domain Deepfake Detection with Deep Information Decomposition}

\author{
Shanmin Yang\textsuperscript{\rm 1}\thanks{\textsuperscript{\rm 1} Chengdu University of Information Technology},
Hui Guo\textsuperscript{\rm
2}\thanks{\textsuperscript{\rm 2} University at Buffalo, SUNY},
Shu Hu\textsuperscript{\rm 3}\thanks{\textsuperscript{\rm 3} Purdue University},
Bin Zhu\textsuperscript{\rm 4}\thanks{\textsuperscript{\rm 4} Microsoft Research Asia},
Ying Fu\textsuperscript{\rm 1},
Siwei Lyu\textsuperscript{\rm 2},
Xi Wu\textsuperscript{\rm {1, $*$}},
Xin Wang\textsuperscript{\rm {5, $*$}}\thanks{\textsuperscript{\rm 5} University at Albany, SUNY}
\thanks{\textsuperscript{${^*}$} Corresponding authors: xi.wu@cuit.edu.cn, xwang56@albany.edu}
}

\markboth{IEEE Transactions on Multimedia} 
{Improving Cross-dataset Deepfake Detection with Deep Information Decomposition}

 \maketitle
 \begin{abstract}
 Deepfake technology poses a significant threat to security and social trust. 
 Although existing detection methods have shown high performance in identifying forgeries within datasets that use the same deepfake techniques for both training and testing, they suffer from sharp performance degradation when faced with cross-dataset scenarios where unseen deepfake techniques are tested.
To address this challenge,
we propose a Deep Information Decomposition (DID) framework to enhance the performance of Cross-dataset Deepfake Detection (CrossDF).
Unlike most existing deepfake detection methods, our framework prioritizes high-level semantic features over specific visual artifacts. 
Specifically, it adaptively decomposes facial features into deepfake-related and irrelevant information, only using the intrinsic deepfake-related information for real/fake discrimination. 
Moreover, it optimizes these two kinds of information to be independent with a de-correlation learning module,
thereby enhancing the  model's robustness against various irrelevant information changes and
generalization ability to unseen forgery methods.
Our extensive experimental evaluation and comparison with existing state-of-the-art detection methods validate the effectiveness and superiority of the DID framework on cross-dataset deepfake detection.
\end{abstract}
  

\section{Introduction}

Significant progress in deep learning and generative techniques such as Face2Face \cite{thies2016face2face},
DeepFake \cite{rossler2019faceforensics++}, 
GANs \cite{pumarola2018ganimation,karras2019style}, etc., has enabled the creation of highly realistic face images. 
However, the widespread use of these deepfakes has posed a significant threat to security and social trust. Therefore, it is crucial to develop effective methods to detect face forgery.

\begin{figure}[t] 
\centering
\includegraphics[width=1.0\linewidth]{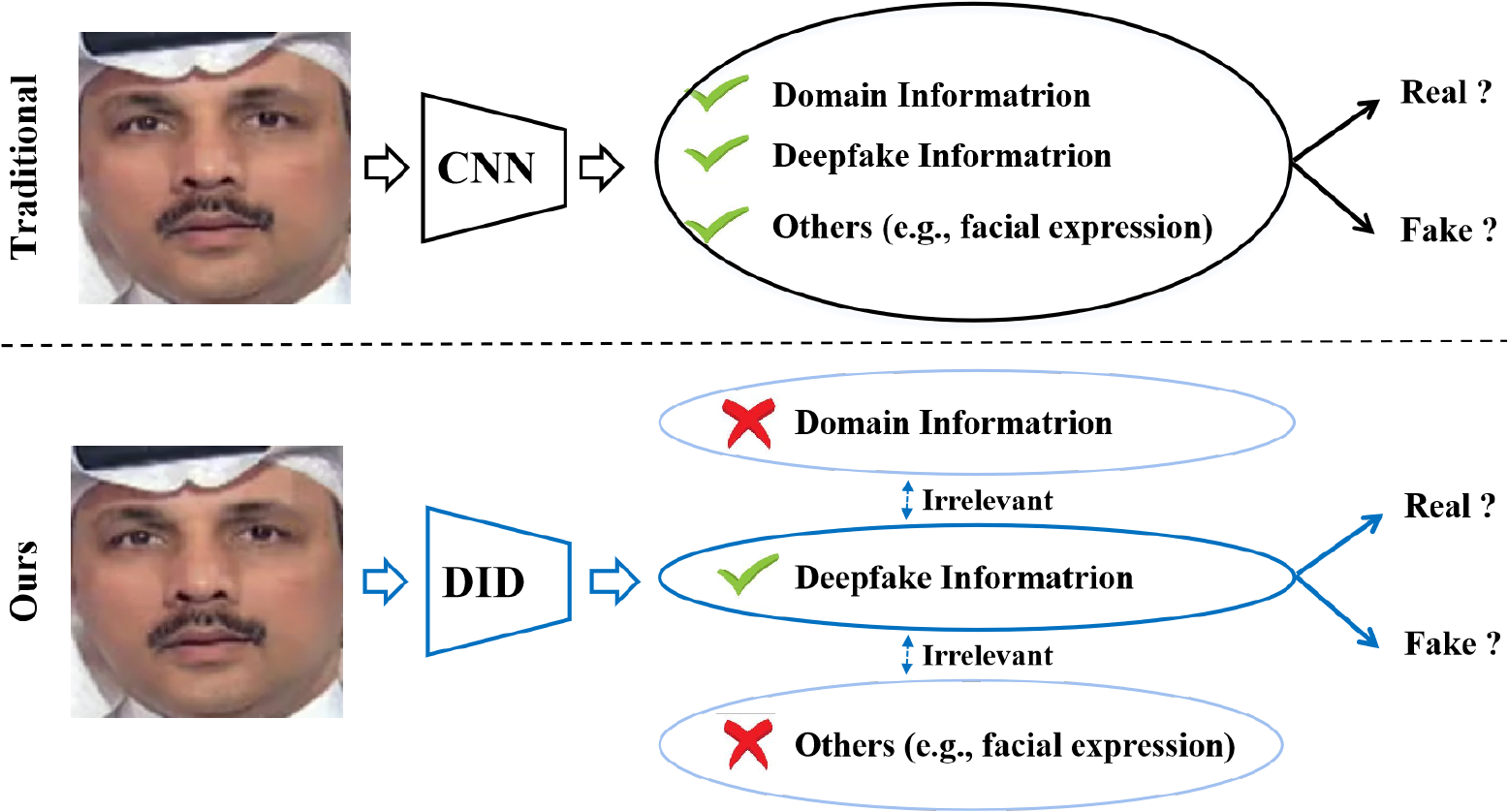}
\caption{Various information changes entangled with the deepfake information in traditional methods (\textbf{top})
would affect real/fake classification accuracy, leading to a sharp degradation in performance when the discrepancies in these components between the training and test sets are more significant than the differences between real and deepfake information.
Our deep information decomposition (DID) method (\textbf{bottom}) separates the deepfake information from
various information irrelevant to real/fake classification to improve the robustness 
 of deepfake detection.}
\label{fig:motivation}
\end{figure}

Numerous efforts have been devoted to deepfake detection in recent years.
Most existing works focus on specific visual artifacts resulting from the deepfake creation process,
such as discrepancies across blending boundaries of real and fake faces \cite{li2020face},
differences of head poses \cite{yang2019exposing}, affine face warping artifacts \cite{li2018exposing}, 
eye state \cite{li2018ictu}, frequency differences~\cite{frank2020leveraging, qian2020thinking,li2021frequency}, etc.
Although these methods have achieved promising performance in intra-dataset scenarios where both training and testing face images are created with the same deepfake technique, they are likely to overfit the specific artifacts of the deepfake technique and thus may be ineffective in detecting forged faces created with different deepfake techniques.
For instance, the detection method proposed in \cite{qian2020thinking} achieves an AUC score of 0.98 when both trained and tested on the same FaceForensics++ (FF++) deepfake dataset
\cite{rossler2019faceforensics++}. Its AUC score degrades sharply to 0.65~\cite{nadimpalli2022improving,kim2022generalized} when trained on the FF++ dataset and tested on Celeb-DF \cite{li2020celeb}.

In studying the cross-dataset performance degradation problem, we observe that deepfake detection is a type of fine-grained image classification. 
With advances in deep forgery methods, the differences between authentic and deepfake images are becoming more and more subtle,
even subtler than those between deepfake images synthesized from the same authentic image with different forgery techniques. 
In addition, features extracted from deepfake images by general deep neural networks (e.g., ResNet-50 \cite{he2016deep},
EfficientNet\cite{tan2021efficientnetv2}, etc.) always include
various entangled information, such as forgery technique-related (domain) information and 
others (e.g., facial expressions and identities) as shown in Fig. \ref{fig:motivation}. 
It makes detection performance sensitive to any change in any component, especially the most prominent ones. 

\begin{figure*}[t]
\centering
\includegraphics[width=0.95\linewidth]{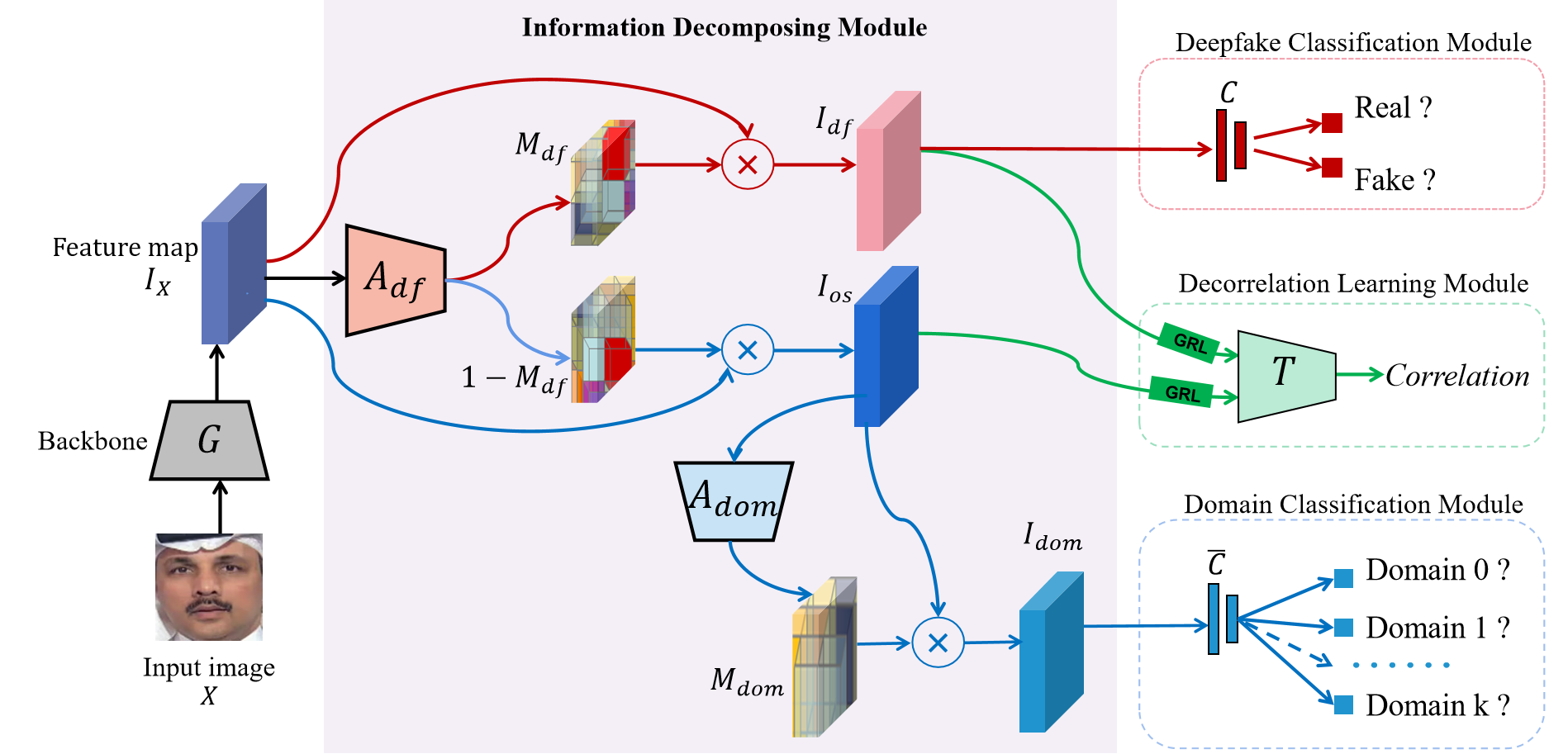}
\caption{Overview of our Deep Information Decomposition (DID) framework.  The feature map $I_X$ of an input face image $X$ from a backbone network $G$ is decomposed into deepfake information $I_{df}$ and non-deepfake information $I_{os}$ adaptively under the guidance of the deepfake attention network $A_{df}$ and the supervision of the deepfake classification module.
The domain attention network $A_{dom}$ and the domain classification module capture the forgery method-related (domain) information $I_{dom}$ 
and ensure that $I_{dom}$ is included in the non-deepfake information but absent in the deepfake information.
In addition, the decorrelation learning module ensures no overlapping between deepfake information and non-deepfake information.
This module consists of an information estimation network $T$, which functions in a max-min manner with the information decomposition module
through the gradient reversal layer (GRL). $C$ and $\overline{C}$ are the deepfake and domain classifiers, respectively.}
\label{fig:framework}
\end{figure*}

Motivated by these observations, we propose a Deep Information Decomposition (DID) framework for cross-dataset deepfake detection,
as shown in Fig. \ref{fig:framework}.
Unlike existing methods, we focus on high-level semantic features rather than low-level deepfake visual traces.
Specifically, we denote face images forged by different deepfake methods as distinct data domains and formulate cross-dataset deepfake detection as a domain generalization problem.
Then, we adaptively decompose the deepfake facial information into deepfake information,  
forgery technique information (e.g., Face2Face \cite{thies2016face2face}, DeepFake \cite{rossler2019faceforensics++}), and others
using two attention modules. 
Only the deepfake information is used for genuine and sham discrimination.
Furthermore, we introduce a de-correlation learning module to promote the deepfake information to be independent of irrelevant information, 
thus improving the detection generalizability to irrelevant variations, including different datasets and forgery methods.
 Extensive experiments demonstrate the effectiveness and superiority of our framework.
 We achieve new state-of-the-art performance on cross-dataset deepfake detection.
 In summary, our main contributions are:
\begin{itemize}
\item We introduce a new end-to-end Deep Information Decomposition (DID) framework that decomposes deepfake face image information into deepfake information and deepfake-irrelevant information.
 By formulating cross-dataset deepfake detection as a domain generalization problem, we enhance the generalization capability of deepfake detectors.
\item A de-correlation learning module is introduced to encourage the independence of the decomposed components
without knowing (assuming) their distribution functions or relationships, which can intrinsically improve the robustness of deepfake detection.
\item We conducted extensive experiments that demonstrated the superiority of our framework, achieving state-of-the-art performance on the challenging cross-dataset deepfake detection task.
\end{itemize}

\section{Related work}
This section provides a brief review of deepfakes, cross-dataset deepfake detection, and information decomposing.
For more details about the deepfake techniques and deepfake datasets, please refer to \cite{nguyen2022deep, Karras_2019_CVPR}.

\subsection{Deepfakes}
Deepfake broadly refers to manipulated or synthetic media (e.g., images, sounds, etc.) that convincingly mimic natural content \cite{zhang2022deepfake, nguyen2022deep}. In this paper, we focus on deepfake faces. We classify existing deepfakes into two types, transfer-based and synthesis-based deepfake methods.

A transfer-based deepfake method tries to manipulate target faces by replacing the faces or facial attributes (e.g., expression, mouth, eye, etc.)  with reference faces. For example, 
Face2Face \cite{thies2016face2face} reenacts the person in a target video with expressions of another person while preserving the identity of the target face. 
FaceSwap \cite{Faceswap}  and DeepFake \cite{rossler2019faceforensics++} replace the face region of a target video with that of a reference source video.
Neural Textures \cite{thies2019deferred} combines learnable neural textures (from the reference video)  with deferred neural rendering to
manipulate facial expressions corresponding to the mouth region.
The method in \cite{bao2018towards} decouples a face image into identity and attributes, then generates a new identity-preserving face image by recombining this identity with attribute features from a different face image.
Almost all deepfake methods of this type require reference faces 
and pay attention to blending manipulated and unmanipulated parts to improve the realism of the deepfake.

A synthesis-based method synthesizes non-existing faces or face attributes (e.g., skin color, hair color, etc.) without using any reference face.  
Generative adversarial networks (GANs) and 3D Morphable Models (3DMM) are popularly used in this type of method. 
For instance, the method in \cite{geng20193d}  proposes a 3DMM-guided way to synthesize arbitrary expressions while preserving face identity.
StyleGAN~\cite{Karras_2019_CVPR} generates a deepfake dataset with high variety and quality based on a style-based generator.
GANprintR~\cite{neves2020ganprintr} is a GAN-based method designed to generate entirely realistic deepfake faces.

\subsection{Cross-Dataset Deepfake Detection}

With the development of deepfake methods, many deepfake detection methods \cite{zhao2021multi,dong2022protecting,shiohara2022detecting,dong2022explaining}
have also been proposed. They can achieve satisfying performance when testing and training face images are 
within the same deepfake dataset but suffer from sharp performance degradation when tested on datasets 
generated with different deepfake techniques and/or types than those in the training datasets.
To address this issue, some works suggest data augmentation to simulate unseen data. For example,
Zhao et al. \cite{zhao2021learning} propose dynamic data augmentation methods to generate new data.
Nadimpalli et al. \cite{nadimpalli2022improving} use a reinforcement learning-based image augmentation model to reduce the shift of the cross dataset (domain).
Several works focus on the distribution of different forgery datasets. For example,
Yu et al.~\cite{yu2022improving} aim to capture common forgery features over different forgery datasets.
Kim et al.~\cite{kim2022generalized} distinguish deepfakes according to color-distribution changes that appeared in the face-synthesis process.
Yu et al.~\cite{yu2023narrowing} propose to narrow the distribution gaps across various forgery types via employing Adaptive Batch and Instance Normalization, generating bridging samples, and performing cross-domain alignment, 
thereby enhancing the model's ability to discern unseen types of fake faces. Yin et al.~\cite{yin2024improving} improve the generalization of deepfake detection across different domains by minimizing invariant risk 
through a learning paradigm that focuses on critical domain-invariant features and aligned representations.
Huang et al.~\cite{huang2023anti} propose a video-level contrastive learning method that maintains a closer distance within data under different compression
levels, thus improving the performance in detecting deepfakes with different compression levels.
However, cross-dataset deepfake detection is still a challenging and critical issue within the realm of deepfake detection,

\subsection{Information Decomposing}

 Information decomposing, which aims to decouple complex, entangled information into distinct semantic components and extract those relevant to specific tasks, has been widely applied across various computer vision applications. 
 The methods in~\cite{tran2017disentangled, wang2019decorrelated} separate identity-dependent information from pose and age variations, respectively, 
to reduce the influence of pose/age discrepancy on face recognition. The method in~\cite{wu2019disentangled} disentangles facial representations into identity and modality information for NIR-VIS heterogeneous face recognition. 
For deepfake detection, Hu et al.~\cite{hu2021improving} identify forgery-related regions through feature disentanglement and train the forgery detector using these regions across different scales.
Liang et al.~\cite{liang2022exploring} separate artifact features from content information to reduce the interference of content information to forgery detection.
Yu et al.~\cite{yu2024fdml} design a framework for face forgery detection that separates forgery-relevant features from source-relevant features progressively from image-level to feature-level through feature disentangling and multi-view learning. 
Yan et al.~\cite{yan2023ucf} use a multi-task learning strategy and a conditional decoder to separate common forgery features relevant to deepfake detection from those that are either irrelevant or method-specific.

In this paper, we present an information decomposition framework that achieves disentanglement via a complementary attention mechanism, 
unlike those in~\cite{hu2021improving, liang2022exploring} that realize information disentanglement via feature encoding and decoding with three elaborately designed reconstruction losses (i.e., self-reconstruction,  cross-reconstruction, and feature reconstruction).
In addition, we introduce a deep decorrelation module to encourage the separated forgery-relevant features used for deepfake detection to be independent of other features, which are ignored in some works, such as \cite{yu2024fdml}, 
thereby enhancing the model's robustness and generalization ability.

\section{Our Method}

The pipeline of our proposed method is shown in Fig.~\ref{fig:framework}. Specifically, for an input image $X$, we use a CNN-based feature extractor $G$ parameterized by $\theta$ to extract its representative features, which can be represented as $I_X:=G(\theta; X)$. Then we decompose these features into three parts: the deepfake-related representation, $I_{df}$, which is the main information used for detecting deepfakes; 
the domain-related representation, $I_{dom}$, which can be used to track the associated forgery technique or method that generates it; and the remainder representation. The information decorrelation module optimizes the decoupled deepfake information $I_{df}$ to be independent of the other representations, thereby enhancing the decomposition performance. The robust deepfake classification module is designed to learn a model that can effectively classify deepfakes in the presence of imbalanced datasets, thus encouraging its generalization ability. 
The domain classification module is designed to identify the domain to which $I_{dom}$ belongs.
Before describing these modules, we introduce the commonly used notation.

\subsection{Notation}

Our method takes images from existing deepfake datasets as input data. Let $\mathcal{S}=\{(X_i, Y_i, D_i)\}_{i=1}^n$ be a training dataset that contains images $X_i\in \mathbb{R}^d$ and their corresponding labels $Y_i\in\{0,1\}$, where 0 denotes real and 1 indicates fake. $D_i:=[D_i^0,D_i^1,...,D_i^k]^\top$ represents the domain label of $X_i$, where the domain size of fake data $k\geq 1$ and $D_i^j\in\{0,1\}, \forall j\in\{0,1,...,k\}$. In particular, $D_i^j=1$ indicates that $X_i$ is from the $j$-th domain. Specifically, $X_i$ is from the real data domain if $j=0$ and from the fake data domain $j$ (i.e., forged by the method $j$) if $j>0$. For example, the fake images in the FF++ dataset {\cite{rossler2019faceforensics++}} are generated by four face manipulation methods: Deepfakes {\cite{Faceswap}}, Face2Face {\cite{thies2016face2face}}, FaceSwap {\cite{Faceswap}}, and NeuralTextures {\cite{thies2019deferred}}. Therefore, $k=4$. In this work, we assume that each $X_i$ comes from only one domain.

\subsection{Information Decomposition Module}

Motivated by \cite{2021Heterogeneous}, the information decomposition module consists of a deepfake attention network $A_{df}$ parameterized by $\psi$ (denoted as $A_{df}(\psi;.)$)
and a domain attention network $A_{dom}$ parameterized by $\varphi$ (denoted as $A_{dom}(\varphi;.)$), 
as shown in Fig. \ref{fig:framework}. Taking the face information $I_X$ embedded with entangled information as input, 
the deepfake attention network focuses on deepfake-relevant information, thereby it
decomposes $I_X$ into two complementary components: 
the deepfake-relevant information $I_{df}$ and the deepfake-irrelevant information $I_{os}$. This process can be formulated as follows,
\begin{align*}
&M_{df}=A_{df}(\psi;I_X),   \\ \notag
&I_{df}=M_{df} \otimes I_X,   \\ \notag
&I_{os}=(1-M_{df}) \otimes  I_X, 
\end{align*}
\noindent
where $M_{df}\in {[0, 1]}^{c\times h\times w}$ is the deepfake-relevant information attention map; 
$\otimes$ represents the Hadamard product.

After receiving the deepfake-irrelevant information $I_{os}$, the domain attention network $A_{dom}$ 
focuses on extracting and modeling explicitly forgery technique information. It decomposes the deepfake-irrelevant information $I_{os}$ 
into the forgery technique-related information $I_{dom}$ and others as follows, 
\begin{align*}      
&M_{dom}=A_{dom}(\varphi;I_{os}),\\  
&I_{dom}=M_{dom}  \otimes  I_{os},
\end{align*}
\noindent
where $M_{dom}\in {[0, 1]}^{c\times h\times w}$ is the forgery technique-related information attention map.

\begin{figure*}[t]
\centering
\includegraphics[width=0.8\linewidth]{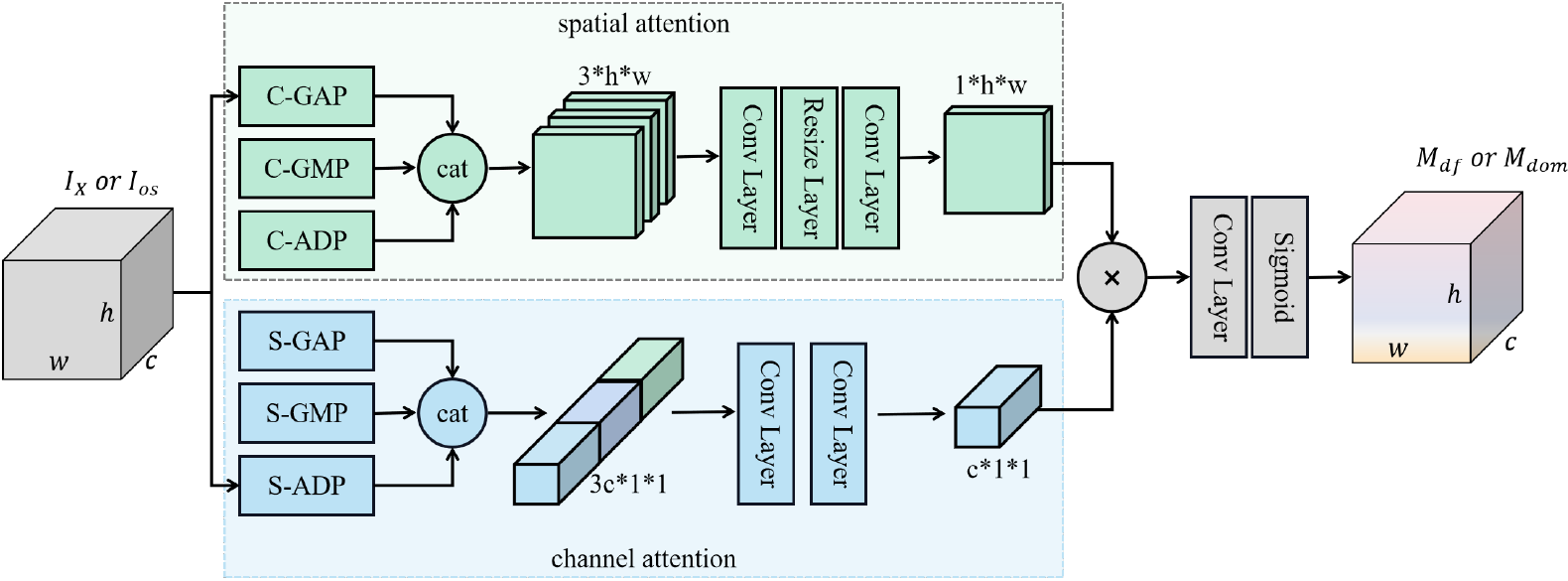}
\caption{Architecture of the deepfake (domain) attention network. This network takes the face information $I_X$ (deepfake-irrelevant information $I_{os}$) as input, and then learns to produce an attention map that highlights the significance (potential) of the input data being correlated with deepfake-relevant (domain-relevant) information. 
``cat" means concatenating all input data along the channel dimension; $\otimes$ represents the Hadamard product; 
``c-" represents cross-channel; ``s-" represents cross-spatial; 
``GAP", ``GMP", and ``ADP" are global average pooling, global max pooling, and adaptive pooling, respectively.}
\label{attention_net}
\end{figure*}

The deepfake attention network $A_{df}$ and the domain attention network $A_{dom}$ are designed to simultaneously learn attention maps from the spatial and channel dimensions. They share the same network architecture as shown in Fig. \ref{attention_net}, where each convolution (Conv) Layer is followed by a PReLU activation function.  S-ADP is implemented by a channel-wise spatial convolution layer followed by a sum pooling layer, while C-ADP is implemented with a 1×1 convolution layer.
Both $A_{df}$ and $A_{dom}$ are optimized to capture the significant deepfake-relevant and domain-relevant information within the input data, respectively.

\subsection{Decorrelation Learning Module}

The disentangled components (deepfake and non-deepfake information) are expected to be distributed in two distinct parts. To this end, orthogonal constraints are generally employed on these disentangled components \cite{he2017learning,he2018wasserstein,wang2019decorrelated}. 
However, linear dependence/independence can hardly characterize the complex relationships between deepfake information and non-deepfake information in a high-dimensional and nonlinear space.
In contrast, mutual information \cite{kinney2014equitability} (MI) is capable of capturing arbitrary dependencies between any two variables.

With this motivation, we apply mutual information to evaluate dependencies between deepfake information $I_{df}$
and non-deepfake information $I_{os}$, formulated as follows:
\begin{equation}
\label{MI_define}
\text{MI}(I_{df}; I_{os}) = \mathbb{D}_{\text{KL}}(P(I_{df},I_{os})|| P(I_{df}) \otimes P(I_{os})),
\end{equation}
where $P(\cdot,\cdot)$ is the joint probability distribution, $P(\cdot)$ denotes the marginal probability distribution, and $\mathbb{D}_{\text{KL}}$ is the Kullback–Leibler divergence {\cite{joyce2011kullback}}.

Since the probability densities $P(I_{df},I_{os})$ and $P(I_{df}) \otimes P(I_{os})$ are unknown, 
it is difficult to directly minimize $\text{MI}(I_{df}; I_{os})$. 
Belghazi et al. \cite{belghazi2018mutual} pioneer a Mutual Information Neural Estimation (MINE) to the lower bound of MI's Donsker-Varadhan representation.
Then \cite{hjelm2018learning} advises a Jensen-Shannon MI estimator (based on the Jensen-Shannon divergence \cite{menendez1997jensen}), which was shown to be more stable and provided better results.

Inspired by \cite{hjelm2018learning}, we construct a mutual information estimation network $T$ with parameterizes $\phi$ to approximate $\text{MI}(I_{df}; I_{os})$ as follows,
\begin{align*}
\text{MI}(I_{df}; I_{os}) &\geq \hat{I}^{JSD}(I_{df}; I_{os}) \\
&= \mathbb{E}_{x \sim P\left(I_{df}, I_{os}\right)}\left[\log \sigma(T(\phi;x))\right] \\ \notag 
& +\mathbb{E}_{x \sim P\left(I_{df}\right) \otimes P\left(I_{os}\right)} \left[\log \left(1\!-\!
 \sigma(T(\phi;x))\right)\right],
\end{align*}
where $\sigma$ is the sigmoid function; $T(\phi;\cdot):\mathbb{R}^{d_x}\rightarrow \mathbb{R}$ acts as the discriminator function in GANs ($d_x$ is the dimension of $I_{df}$ and $I_{os}$), it aims to estimate and maximize the lower bound of $\text{MI}(I_{df}; I_{os})$,
while the target of the previously designed information  
decomposition module (acting as the generator function in GANs) is to minimize the MI value between $I_{df}$ and $I_{os}$ to achieve a sufficient separation.
Specifically, we have the following learning objectives:
\begin{align*}
\label{decorrleation}
\mathcal{L}_{dec} =\min _{\theta, \psi} \max _{\phi}&(\mathbb{E}_{x \sim P\left(I_{df}, I_{os}\right)}\left[\log \sigma(T(\phi;x))\right] \\ \notag 
 +&\mathbb{E}_{x \sim P\left(I_{df}\right) \otimes P\left(I_{os}\right)} \left[\log \left(1\!-\!
 \sigma(T(\phi;x))\right)\right]).
\end{align*}

To implement the aforementioned min-max game using standard back-propagation (BP) training, we add a Gradient Reversal Layer (GRL) \cite{ganin2015unsupervised} before the network $T$ (shown in Fig. \ref{fig:framework}). 
In the back-propagation procedure, GRL transmits the gradient by multiplying a negative scalar, $-\beta$, from the subsequent layer to the preceding layer, where we set $\beta\in(0,1)$ in practice. This trick is also used in several existing works such as \cite{belghazi2018mutual, hjelm2018learning}.  
The network $T$ consists of three convolution layers (a ReLU activation follows each layer) and a fully connected layer (FC).

\subsection{Robust Deepfake Classification Module}

After we obtain deepfake-related information $I_{df}$, we need to consider how to use it to learn a deepfake detection model. In the literature, the Binary Cross-entropy (BCE) loss is widely used to train a deepfake detection model. However, it is well-known that the BCE loss is not robust to the imbalance data, especially for deepfake datasets.
Using BCE loss to train models on a certain deepfake dataset may cause significant performance degradation when testing on another deepfake dataset \cite{pu2022learning}.

With this observation, we propose a robust deepfake detection loss to enhance the generalization ability of the trained model using deepfake-related information $I_{df}$ instead of complete information $I_X$. Our loss is inspired by the AUC metric since it is a robust measure to evaluate the classification capability of a model, especially when facing imbalanced data. Specifically, it estimates the size of the area under the receiver operating characteristic (ROC) curve (AUC) \cite{he2009learning}, which is composed of False Positive Rates (FPRs) and True Positive Rates (TPRs). However, the AUC metric cannot be directly used as a loss function since it is challenging to compute during each training iteration. Inspired by \cite{pu2022learning}, we use the normalized WMW statistic \cite{yan2003optimizing}, equivalent to AUC, to design our loss function.

\begin{algorithm}[t]
\caption{Deep Information Decompositio}
\label{alg:algorithm1}
\SetAlgoLined
\KwIn{A training dataset $\mathcal{S}$ of size $n$, $\beta$, max\_iterations,  num\_batch, $\eta_\theta$, $\eta_\psi$, $\eta_\varphi$, $\eta_\phi$, $\eta_\omega$, and $\eta_{\overline{\omega}}$ }
\KwOut{A robust Deepfake detector with parameters $\theta^*$, $\psi^*$, and $\omega^*$}
\textbf{Initialization: $\theta_0$, $\psi_0$, $\varphi_0$, $\phi_0$, $\omega_0$, $\overline{\omega}_0$, $l=0$}

\For{$e=1$ to \emph{max\_iterations}}{
\For{$b=1$ to \emph{num\_batch}}{
\mbox{Sample a mini-batch $\mathcal{S}_b$ from $\mathcal{S}$}

Update parameters with (\ref{eq:update_rule}).

$l \leftarrow l+1$
}
}
\Return{$\theta^* \leftarrow \theta_{l}$, $\psi^* \leftarrow \psi_{l}$, $\omega^* \leftarrow \omega_{l}$}
\end{algorithm}

Specifically, we define a set of indices of fake instances and real instances as $\mathcal{F}=\{i|Y_i=1\}$ and $\mathcal{R}=\{i|Y_i=0\}$, respectively. We add a multilayer perceptron (MLP) $C:\mathbb{R}^{d_x}\rightarrow \mathbb{R}$ ($d_x$ is the dimension of $I_{df}$) parameterized by $\omega$ to distinguish fake and real instances, where the input is $I_{df}$ and the output is a real value. 
Network $C$ predicts input $I_{df}$ to be fake with probability $\sigma(C(\omega;I_{df}))$. Without loss of generality, $C(\omega;I_{df})$ induces the prediction rule such that the predicted label of $I_{df}$ can be $\mathbb{I}[\sigma(C(\omega;I_{df}))\geq 0.5]$, where $\mathbb{I}[\cdot]$ is an indicator function with $\mathbb{I}[a]=1$ if a is true and 0 otherwise. For simplicity, we assume $C(\omega;I_{df}^{X_i})\neq C(\omega;I_{df}^{X_j})$ for any $X_i\neq X_j$ (ties can be broken in any consistent way), where $I_{df}^{X_i}$ represents the deepfake information of the sample $X_i$. Then the normalized WMW can be formulated as follows,
\begin{equation*}
    \begin{aligned}
    \text{WMW}=\frac{1}{|\mathcal{F}||\mathcal{R}|}\sum_{i\in \mathcal{F}}\sum_{j\in \mathcal{R}}\mathbb{I}[C(\omega;I_{df}^{X_i})>C(\omega;I_{df}^{X_j})],
    \end{aligned}
\end{equation*}
where $|\mathcal{F}|$ and $|\mathcal{R}|$ are the cardinality of $\mathcal{F}$ and $\mathcal{R}$, respectively. However, WMW is non-differentiable due to the indicator function, which is the main obstacle to using it as a loss. Therefore, we use its alternative version \cite{yan2003optimizing}:
\begin{equation*}
    \begin{aligned}
    \mathcal{L}_{\text{AUC}}=\frac{1}{|\mathcal{F}||\mathcal{R}|}\sum_{i\in \mathcal{F}}\sum_{j\in \mathcal{R}}E(C(\omega;I_{df}^{X_i}),C(\omega;I_{df}^{X_j})),
    \end{aligned}
\end{equation*}
with 
\begin{equation} 
  \resizebox{\linewidth}{!}{$
   \begin{aligned}
    &E(C(\omega;I_{df}^{X_i}),C(\omega;I_{df}^{X_j}))\\
    &:=\!\!\left\{\begin{matrix}
(-(C(\omega;I_{df}^{X_i})\!-\!C(\omega;I_{df}^{X_j})\!-\!\gamma))^p, & \!\!\!\!\!\!C(\omega;I_{df}^{X_i})-C(\omega;I_{df}^{X_j})<\gamma,\\ 
0, & \text{otherwise},
\end{matrix}\right.
    \end{aligned} 
\label{AUC loss def} $}
\end{equation}
where $0<\gamma\leq 1$ and $p>1$ are two hyperparameters. We combine this AUC loss and the conventional BCE loss $\mathcal{L}_{\text{BCE}}:= -\frac{1}{n}\sum_{i=1}^n [Y_i\cdot \log(\sigma(C(\omega; I_{df}^{X_i})))+(1-Y_i)\cdot \log(1-\sigma(C(\omega; I_{df}^{X_i})))]$ to construct a learning objective for robust deepfake classification:
\begin{equation}
    \begin{aligned}
    \mathcal{L}_{cls}= \alpha \mathcal{L}_{\text{BCE}}+ (1-\alpha) \mathcal{L}_{\text{AUC}}
    \end{aligned}
\label{cls loss}
\end{equation}
where $\alpha$ is a hyperparameter designed to balance the weights of the BCE loss and the AUC loss.

\subsection{Domain Classification Module}

A domain classification module is also designed using another MLP $\overline{C}:\mathbb{R}^{d_{I_X}}\rightarrow \mathbb{R}^{k+1}$ parameterized by $\overline{\omega}$ to map the forgery method related-domain information  $I_{dom}$ into a (k+1)-dimensional domain vector. Specifically, we have $\overline{C}(\overline{\omega};I_{dom})=[\overline{C}^0(\overline{\omega};I_{dom}), \overline{C}^1(\overline{\omega};I_{dom}), ...,\overline{C}^k(\overline{\omega};I_{dom})]^\top$, where $\overline{C}^j(\overline{\omega};I_{dom})$ is the $j$-th domain prediction.
We then apply the softmax function to compute the probability of each domain that $I_{dom}$ belongs to and combine its domain label to construct a domain classification loss based on the cross-entropy (CE) loss. Therefore, we have
\begin{align*}
\label{class_dom}
\mathcal{L}_{dom}= -\frac{1}{n}\sum_{i=1}^n\sum_{j=0}^k D_i^j\log(S[\overline{C}^j(\overline{\omega};I_{dom}^{X_i})])
\end{align*}
where $S[\overline{C}^j(\overline{\omega};I_{dom}^{X_i})]\in(0,1)$ is the $j$-th domain predicted probability for the domain information of $I_{X_i}$ after using softmax operator $S[\cdot]$.

\noindent
\textbf{Overall Loss}.
To sum up, the proposed framework is optimized with the following final loss function:
\begin{equation}
\label{lossfc}
\mathcal{L}= \lambda_{dec} \mathcal{L}_{dec} + \lambda_{cls} \mathcal{L}_{cls} + \lambda_{dom} \mathcal{L}_{dom}
\end{equation}
where $\lambda_{dec}$, $\lambda_{cls} $, and $\lambda_{dom}$ are hyperparameters that can balance these loss terms. In practice, the optimization problem in (\ref{lossfc}) can be solved with an iterative stochastic gradient descent and ascent approach \cite{beznosikov2022stochastic}. Specifically, we first initialize the model parameters $\theta$, $\psi$, $\varphi$, $\phi$, $\omega$, and $\overline{\omega}$. 
Then we alternate uniformly at random a mini-batch $\mathcal{S}_b$ of training samples from the training set $\mathcal{S}$ and do the following steps on $\mathcal{S}_b$ for each iteration:

\begin{equation}
    \begin{aligned}
    \begin{pmatrix}
\theta_{l+1}\\ 
\psi_{l+1}\\ 
\varphi_{l+1}\\ 
\phi_{l+1}\\ 
\omega_{l+1}\\ 
\overline{\omega}_{l+1}\\
\end{pmatrix}=\begin{pmatrix}
\theta_{l}\\ 
\psi_{l}\\ 
\varphi_{l}\\ 
\phi_{l}\\ 
\omega_{l}\\ 
\overline{\omega}_{l}\\
\end{pmatrix}-
\begin{pmatrix}
\eta_\theta\partial_{\theta} \mathcal{L}|_{\theta=\theta_l}\\ 
\eta_\psi\partial_{\psi} \mathcal{L}|_{\psi=\psi_l}\\ 
\eta_\varphi\partial_{\varphi} \mathcal{L}|_{\varphi=\varphi_l}\\ 
-\eta_\phi \beta\partial_{\phi} \mathcal{L}|_{\phi=\phi_l}\\ 
\eta_\omega\partial_{\omega} \mathcal{L}|_{\omega=\omega_l}\\ 
\eta_{\overline{\omega}}\partial_{\overline{\omega}} \mathcal{L}|_{\overline{\omega}=\overline{\omega}_l}\\
\end{pmatrix},
    \end{aligned}
\label{eq:update_rule}
\end{equation}
where $\mathcal{L}$ is defined on $\mathcal{S}_b$, $\eta_\theta$, $\eta_\psi$, $\eta_\varphi$, $\eta_\phi$, $\eta_\omega$, and $\eta_{\overline{\omega}}$ are learning rates, and $\partial_{\theta} \mathcal{L}$, $\partial_{\psi} \mathcal{L}$, $\partial_{\varphi} \mathcal{L}$, $\partial_{\phi} \mathcal{L}$, $\partial_{\omega} \mathcal{L}$, and $\partial_{\overline{\omega}} \mathcal{L}$ are the (sub)gradient of $\mathcal{L}$ with respect to $\theta$, $\psi$, $\varphi$, $\phi$, $\omega$, and $\overline{\omega}$. 
In the testing phase, we only use the feature extractor $G$, attention module $A_{df}$, and the deepfake classification module $C$. The pseudocode is shown in Algorithm \ref{alg:algorithm1}.


\section{Experiments}

This section evaluates the effectiveness of the proposed framework (i.e., DID) in terms of cross-dataset deepfake detection performance. In the following discussion, we will exchange the ``method" or ``framework" used for DID.

\subsection{Experimental Settings}

\noindent \textbf{Datasets.}
For fair comparisons with the state-of-the-art methods, the two most popular 
FF++ \cite{rossler2019faceforensics++} and Celeb-DF \cite{li2020celeb} datasets are adopted in our experiments. 
In particular, the high-quality (HQ, with a constant compression rate factor of 23) version of FF++ is used in all of our experiments, which contains one real video subset and four fake video subsets, generated using FaceSwap, DeepFakes, Face2Face, and Neural Textures techniques, respectively. Each subset contains 1000 videos, in which 720/140/140 videos are used for training /validation/testing, respectively \cite{rossler2019faceforensics++}.
The Celeb-DF \cite{li2020celeb} dataset contains real and fake videos according to 59 celebrities. Following the official protocols in \cite{li2020celeb}, we use the latest version of Celeb-DF named Celeb-DF V2, which contains 590 real celebrity (Celeb-real) videos, 300 real videos downloaded from YouTube (YouTube-real)and 5639 synthesized celebrity (Celeb-synthesis) videos based on Celeb-real.

\noindent \textbf{Compared Methods and Evaluation Metrics.} 
To evaluate the effectiveness of our framework, we compare it with the following 
state-of-the-art (SOTA) frame-level baseline methods: F$^3$-Net~\cite{qian2020thinking},
CFFs~\cite{yu2022improving}, RL~\cite {nadimpalli2022improving}, Multi-task~\cite{nguyen2019multi}, Two Branch~\cite{masi2020two}, 
 MDD~\cite{zhao2021multi}, and NoiseDF \cite{wang2023noise}.
The results of Multi-task and Two Branch are cited from~\cite{nadimpalli2022improving}, 
and the results of MDD are cited from~\cite{yu2022improving}. 
We consider two evaluation metrics, the area under the receiver operating characteristic curve (AUC) and the equal error rate (EER), 
which are widely adopted in previous works.

\noindent\textbf{Implementation Details.}
In our experiments, EfficientNet v2-L \cite{tan2021efficientnetv2}  pre-trained on the ImageNet dataset is adopted as the backbone for feature extraction. 
Face images in all frames are aligned to $224\times224$ using the MTCNN \cite{zhang2016joint} method. Then they are converted into the grayscale from the RGB before sending to the proposed framework.  
The framework is trained with the Adam optimizer with a weight decay of $5e^{-4}$ and a learning rate of $1e^{-5}$. 
We set the learning rate $\eta_\psi$, $\eta_\varphi$, $\eta_\phi$, $\eta_\omega$, and $\eta_{\overline{\omega}}$ in (\ref{eq:update_rule})
 to be 10 times that of $\eta_\theta$ ($\eta_\theta=1e^{-5}$). 
The batch size is 15 and the number of iterations in each epoch is 6000.
We set $\gamma$ and $p$ in (\ref{AUC loss def}) to 0.15 and 2, respectively. We use $\alpha=0.5$ in (\ref{cls loss}).
The hyperparameters $\lambda_{cls}$, $\lambda_{dom}$, and $\lambda_{dec} $ in (\ref{lossfc}) are set to 1, 1, and 0.01, respectively. 
The hyperparameter $\beta$ in (\ref{eq:update_rule}) is adapted to increase from 0 to 1 in the training procedure as
$\beta= 2.0 / (1.0 + e^{-5 p}) - 1.0$, where $p$ is the ratio of the current training epochs to the maximum number of training epochs. All experiments are conducted on two NVIDIA RTX 3080 GPUs, with Pytorch 1.10 and Python 3.6.

\subsection{Intra-dataset Evaluation}
We evaluate the detection performance of our proposed method DID in the intra-dataset situation,
where the training and test sets are from the FF++ dataset and are disjoint. 
Table~\ref{tests} shows the intra-dataset evaluation results and comparison with the baselines. 
We can see that our method achieves 0.970 on AUC, which outperforms the Multi-task~\cite{nguyen2019multi}, Two Branch~\cite{masi2020two}, and NoiseDF \cite{wang2023noise} methods,
and is competitive with the best performance
(0.998 on AUC score achieved by MDD~\cite{zhao2021multi}).

\begin{table}[t]
\caption {Performance comparison with the baselines in both \textbf{intra-dataset} and \textbf{cross-dataset} scenarios. 
The results of Multi-task and Two Branch are cited from~\cite{nadimpalli2022improving},
and the results of MDD are cited from~\cite{yu2022improving}.
}
\label{tests}
\centering
\begin{tabular}{c|c|c}
\hline
\multirow{2}{*}{Methods} & {Intra-dataset}                      & {Cross-dataset}                   \\ 
& AUC \textuparrow                                                   & AUC \textuparrow                    \\ \hline
{Multi-task} \cite{nguyen2019multi}              &  {0.763}           & {0.543}   \\ 
Two Branch \cite{masi2020two}                   & 0.931               & 0.734     \\ 
MDD \cite{zhao2021multi}                        &  \textbf{0.998}     & 0.674     \\ 
RL  \cite {nadimpalli2022improving}             & 0.994               & 0.669     \\ 
F$^3$-Net \cite{qian2020thinking}               & 0.981               & 0.651    \\ 
CFFs \cite{yu2022improving}                     & 0.976               & 0.742     \\  
NoiseDF \cite{wang2023noise}                    &0.940                & 0.759     \\  
FDML \cite{yu2024fdml}                          &0.996                & 0.731    \\
{DID (Ours)}                                   & 0.970              &  {\textbf{0.779}}   \\ \hline
\end{tabular}
\end{table}

\subsection{Cross-dataset Evaluation}
The cross-dataset generalization performance of the proposed method and comparison with the baselines are also shown in Table~\ref{tests}.
All models are first trained on the training set of the FF++ dataset and then tested on the test set of Celeb-DF v2 (unseen during training). 
From the table, it is evident that the performance of all methods significantly degrades in the challenging cross-dataset scenario compared to the intra-dataset scenario. For example, the performance of MDD~\cite{zhao2021multi} declines
from 0.998 to 0.674. In comparison, our DID method exhibits excellent generalization capability, 
achieving superior performance in this cross-dataset scenario. It exceeds the CFFs~\cite{yu2022improving} and NoiseDF \cite{wang2023noise} methods by a margin of 4.99\% (0.779 vs. 0.742) and 2.635\% (0.779 vs. 0.759) respectively in terms of AUC. 
These experimental results validate the effectiveness and superiority of our framework. 

 \begin{table}[t]
 \caption{Cross-dataset deepfake detection performance. 
All models are trained on the same subset of the DFFD dataset and tested on the Celeb-DF v2 dataset.}
\label{DFFD_exp}
\centering
\begin{tabular}{c|c|c}
\hline
Models                     & AUC \textuparrow     & EER \textdownarrow \\ \hline                              
ResNet50 + BCE             & 0.620                & 0.411        \\
\textbf{ResNet50 + DID}    &\textbf{0.727}        & \textbf{0.332}       \\ \hline\hline
EfficientNet-v2-L + BCE    & 0.716                & 0.344   \\   
\textbf{EfficientNet-v2-L + DID}    &\textbf{0.763}      &\textbf{0.302} \\\hline  
\end{tabular}
\end{table}

\begin{table}[t]
\caption{Ablation study by removing the domain attention module $A_{dom}$ (``w/o $A_{dom}$'') or the decorrelation learning module $T$ (``w/o $T$'') from the DID framework.}
\label{com_anaysis}
\centering
\begin{tabular}{c|c c c |c c }
\hline
\multirow{2}{*}{Models} & \multicolumn{3}{c|}{Modules} & \multirow{2}{*}{AUC \textuparrow} & \multirow{2}{*}{EER \textdownarrow}   \\
\cline{2-4}
                & $A_{df}$     & $A_{dom}$     & $T$        &          &\\   \hline
w/o $A_{dom}$   & $\surd$      & $\times$      & $\surd$    &  0.763   & 0.302\\  
w/o $T$         & $\surd$      & $\surd$       & $\times$   &  0.759   & 0.305\\  
DID             & $\surd$      & $\surd$       & $\surd$    &  0.779   & 0.286 \\ \hline
\end{tabular}
\end{table}

\begin{figure}[t] 
\centering
\subfloat[]{\includegraphics[width=0.45\linewidth]{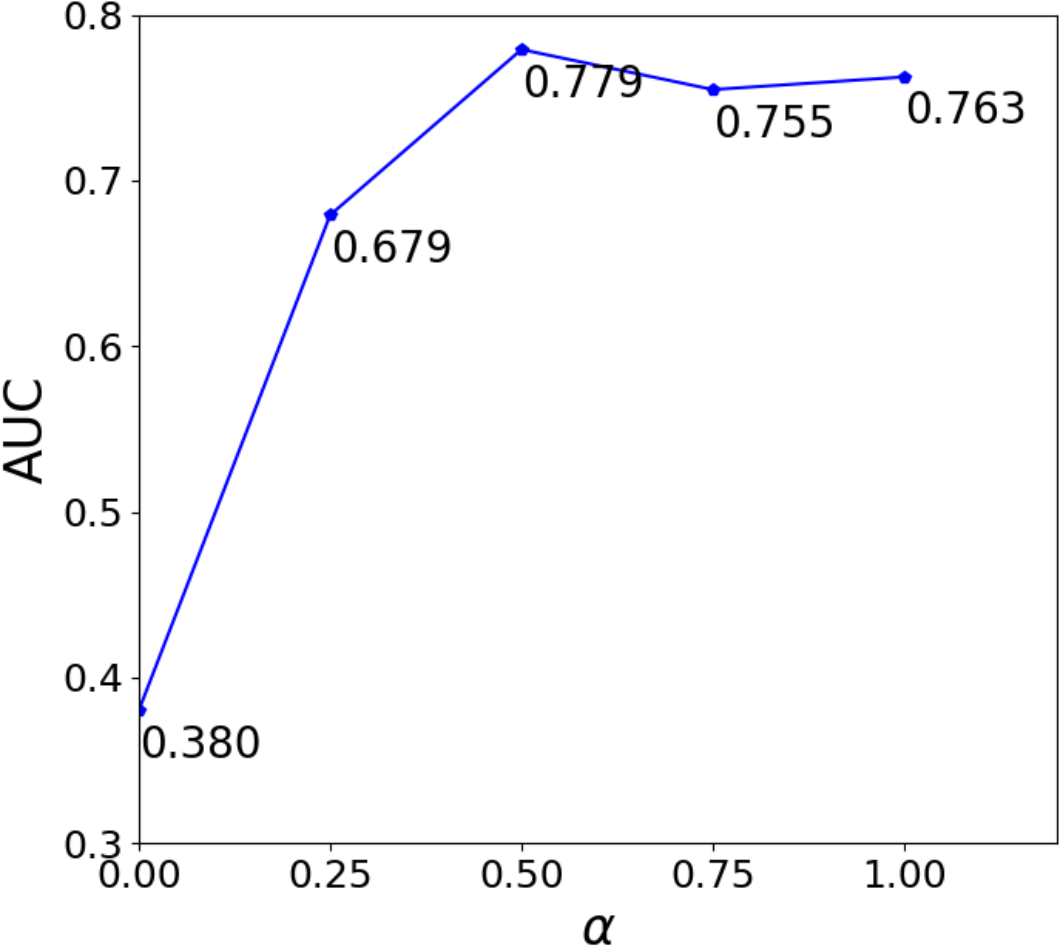}} 
\subfloat[]{\includegraphics[width=0.54\linewidth]{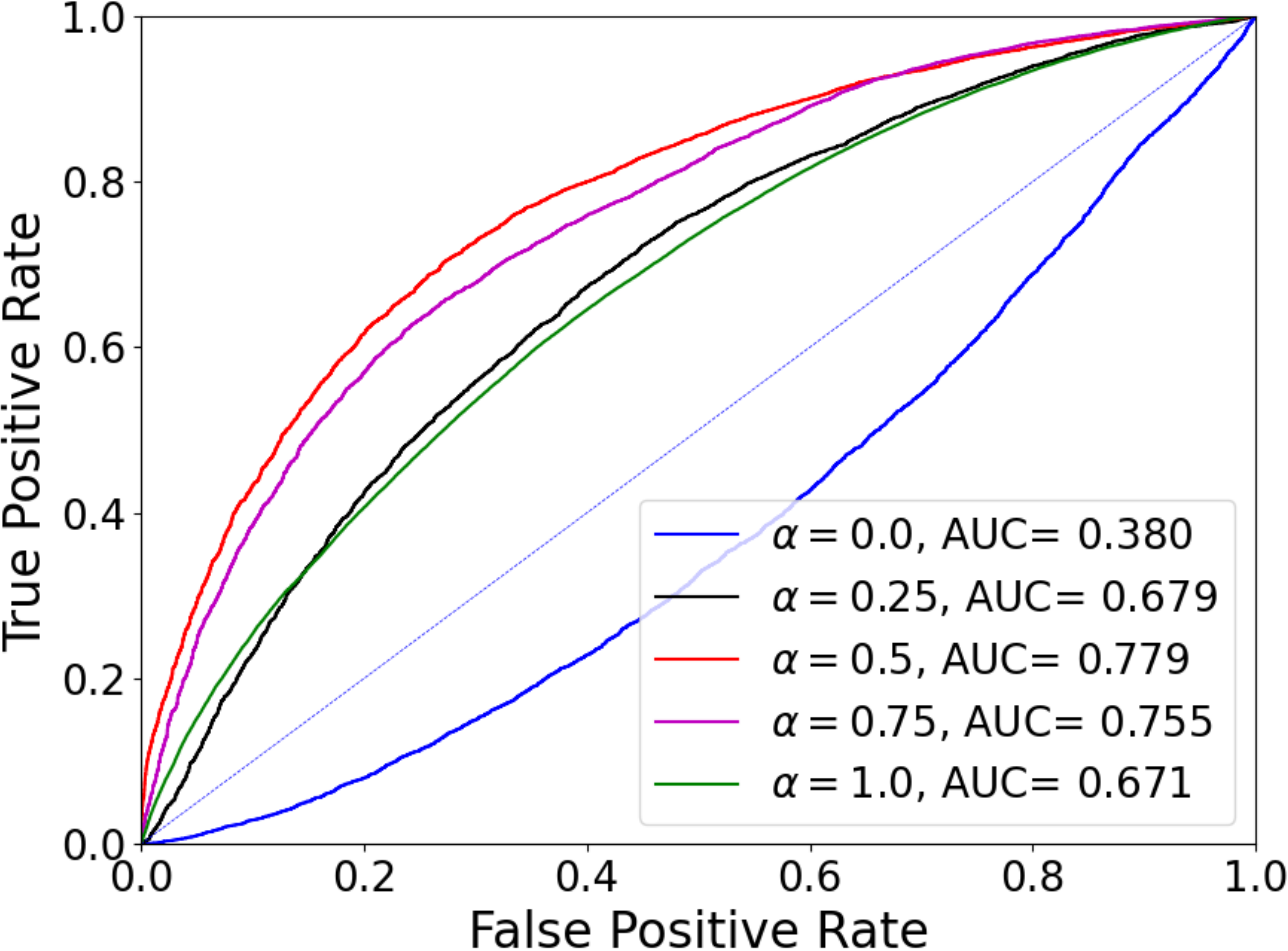}} 
\caption{
Effect of different $\alpha$ values (used as the balance factor between BCE and AUC loss) on the AUC score. 
(a) is AUC with different $\alpha$ values, (b) is ROC with different $\alpha$ values. 
}
\label{alpha-snalysis}
\end{figure}

\begin{figure}[t]
\centering
\includegraphics[width=0.8\linewidth]{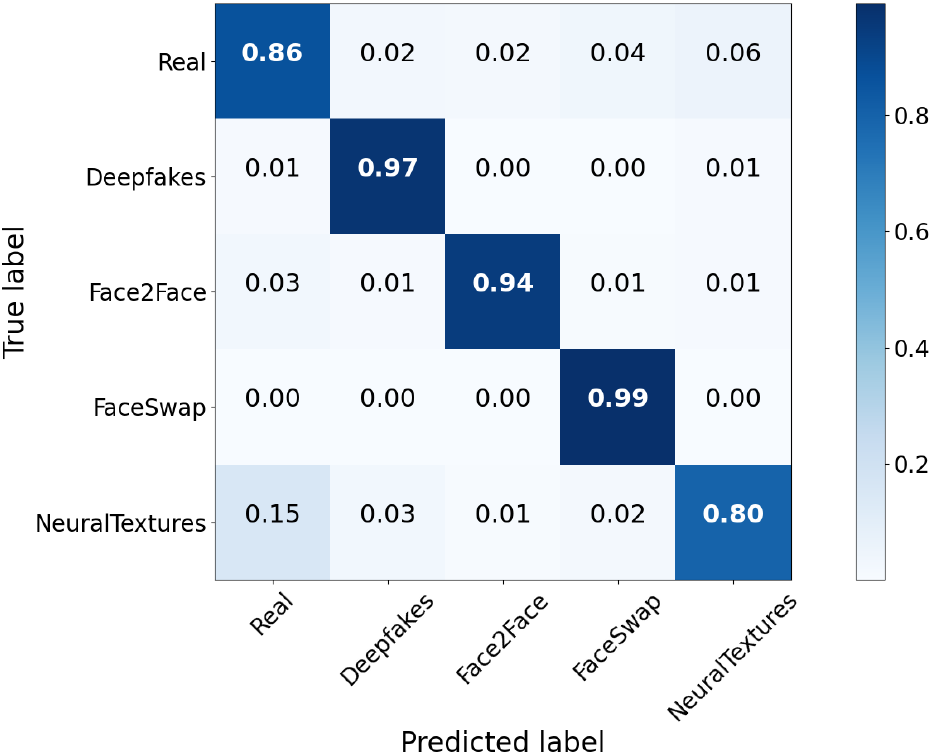}
\caption{Confusion matrix visualization of domain feature classification. Each deepfake technique is recognized by the domain classification module with high accuracy (the value on the diagonal).}
\label{confusion_matrix}
\end{figure}

\noindent\textbf{Effect on Different Training Dataset}
To further demonstrate the applicability of our methods, we train the DID framework on the DFFD dataset~\cite{dang2020detection} 
and evaluate the detection performance on the Celeb-DF dataset.
DFFD is a deepfake dataset composing real images and the corresponding deepfakes created with FaceSwap \cite{Faceswap}, Deepfake, Face2Face \cite{thies2016face2face}, FaceAPP \cite{faceapp}, StarGAN \cite{choi2018stargan}, PGGAN \cite{karras2018progressive} (two versions), and StyleGAN \cite{karras2019style} methods, and
deepfake videos generated by Deep Face Lab \cite{DeepFaceLab}.
The experiments are conducted on a subset of DFFD (excluding deepfake videos that are inaccessible) following the protocols of \cite{dang2020detection}, with different feature extraction backbones. 
Table~\ref{DFFD_exp} shows the experimental results, from which we can see that the proposed DID 
achieves great performance improvement on all backbones fine-tuned with the BCE loss.
Specifically, the improvement relative to the ResNet50 backbone (fine-tuned with the BCE loss) is 17.26\%  (0.727 vs. 0.620) on AUC  and 19.22\% (0.763  vs. 0.716) on EER.
Meanwhile, the improvement relative to the EfficientNet-v2-L backbone is 6.56\% (0.763 vs. 0.716) on AUC 
and 12.21\% (0.302 vs. 0.344) on EER.
These results demonstrate the applicability of our methods on different datasets and different feature extraction backbones      
in cross-dataset deepfake detection.

\subsection{Ablation Study}

\noindent\textbf{The Effect of AUC Loss.}
The impact of hyperparameter $\alpha$ in the AUC loss,  as shown in (\ref{cls loss}), is investigated. Specifically, 
we train our model with different $\alpha\in\{0.0, 0.25, 0.5, 0.75, 1.0\}$ values and show the AUC performance in Fig.~\ref{alpha-snalysis}.
The model trained with only the AUC loss ($\alpha$=0.0) as the deepfake classification loss function gets the lowest AUC of 0.380, 
and the model trained with only the BCE deepfake classification loss ($\alpha$=1.0) obtains an AUC of 0.763.
However, the model trained with an equal weight of the AUC loss and the BCE loss ($\alpha$=0.5) shows the best AUC score of 0.779, which means that the AUC loss can help improve the generalization ability of the model.

\noindent\textbf{The Effect of $A_{dom}$ and $T$ Modules.}
To explore the necessity of the domain attention module $A_{dom}$ and the decorrelation learning module $T$, 
we train the proposed DID framework with either one of the two modules removed. 
The detection results of these models are shown in Table~\ref{com_anaysis}.
We can see from the table that when the domain attention module $A_{dom}$ is removed from DID (``w/o $A_{dom}$'' in Table~\ref{com_anaysis}), the AUC score declines 
by 2.05\%  (from 0.779 to 0.763) and EER increases by 5.59\% (from 0.286 to 0.302) relative to the complete DID version. 
In addition, the model without the decorrelation learning module $T$ (``w/o $T$'' in Table~\ref{com_anaysis}) 
achieves 0.759 on AUC and 0.305 on EER, the performance degradation (relative to DID) is greater than the model w/o $A_{dom}$.
Specifically, the AUC score drops by  2.57\%, and EER increases by 6.64\%.
These results suggest the indispensability of the DID framework's $A_{dom}$ and $T$ modules.

\noindent\textbf{Analysis of Domain Classification Module.}
Fig.~\ref{confusion_matrix}  displays the confusion matrix of the domain feature classification.
We can see that the domain classification module distinguishes all forgery methods with very high accuracy.
Specifically, the average accuracy is 0.91, and the highest accuracy is 0.99
(classification of the FaceSwap method).
These results indicate that domain information is successfully separated from the deepfake features
and captured by the domain classification module,
which aligns with our decomposition objective and proves to be advantageous for deepfake detection.

\begin{figure*}[t] 
\centering
\includegraphics[width=\linewidth]{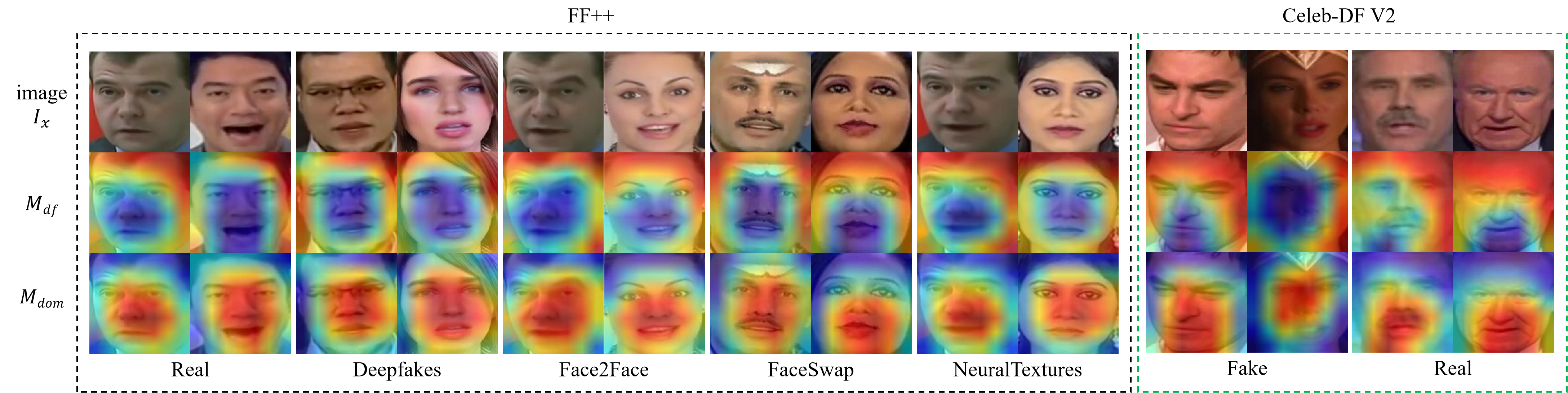}
\caption{Visualization of the real/fake attention map $M_{df}$ and the domain (forgery technique) attention map $M_{dom}$ on the 
FF++ and Celeb-DF v2 datasets. We can see that $M_{dom}$ captures the forgery technique-related information (e.g., the forged region), while 
$M_{df}$ focuses on the information invariant to forgery techniques.
 }
\label{fig:attentionmap}
\end{figure*}

 \begin{figure*}[t]  
\centering
\subfloat[Backbone's deepfake features]{\includegraphics[width=0.3\linewidth]{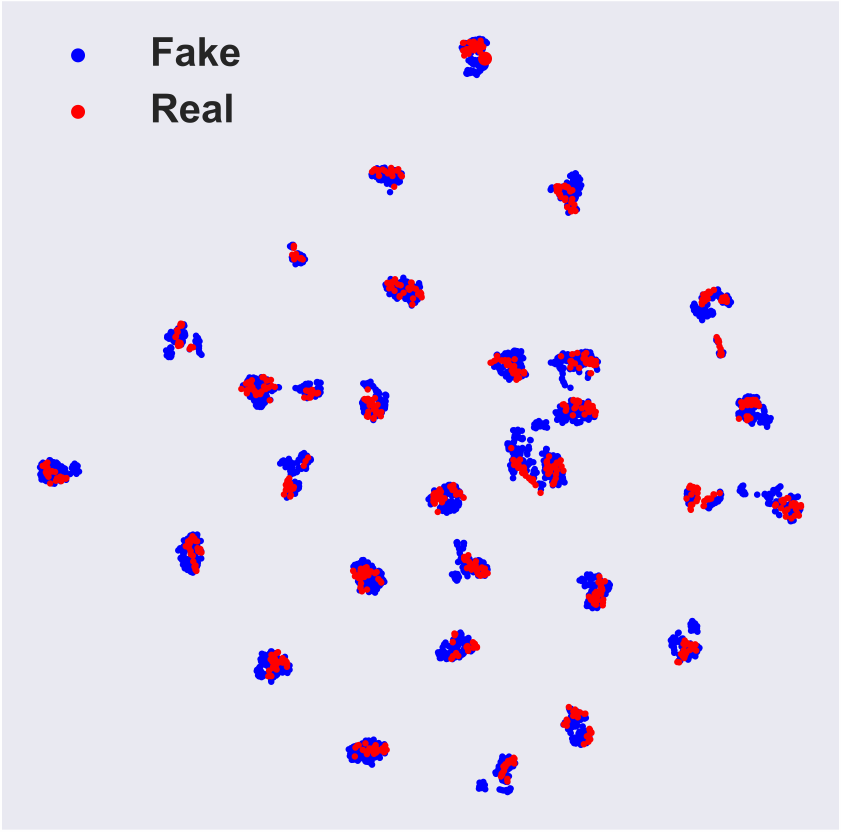}\label{backbone}} 
\hspace{4mm}
\subfloat[DID's deepfake features]{\includegraphics[width=0.3\linewidth]{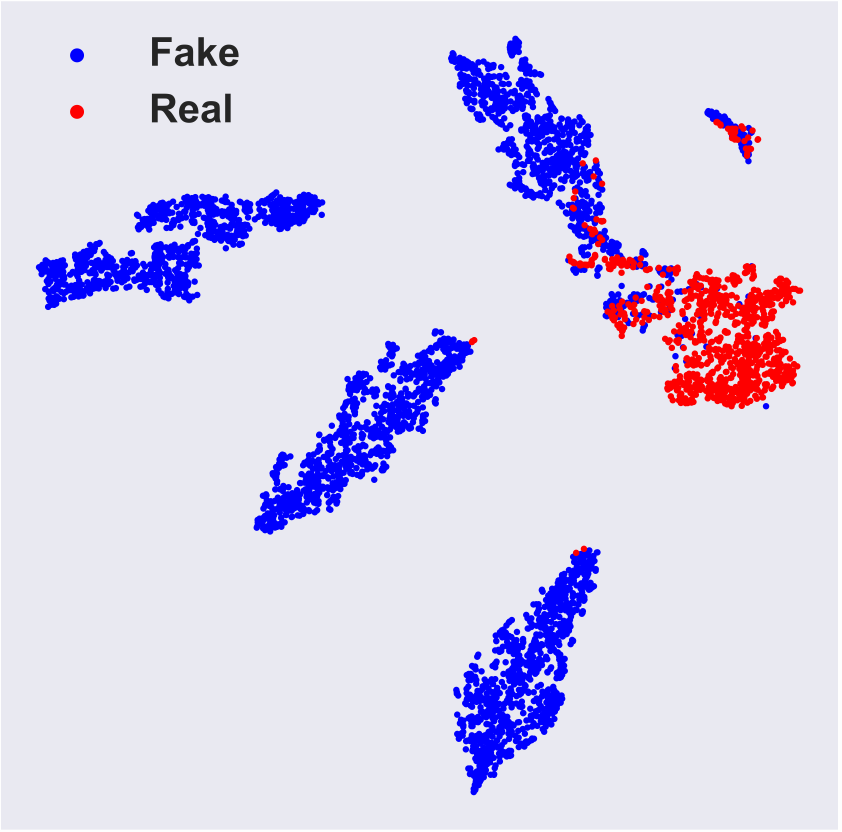}\label{tsne_df}} 
\hspace{4mm}
\subfloat[DID's domain features]{\includegraphics[width=0.34\linewidth]{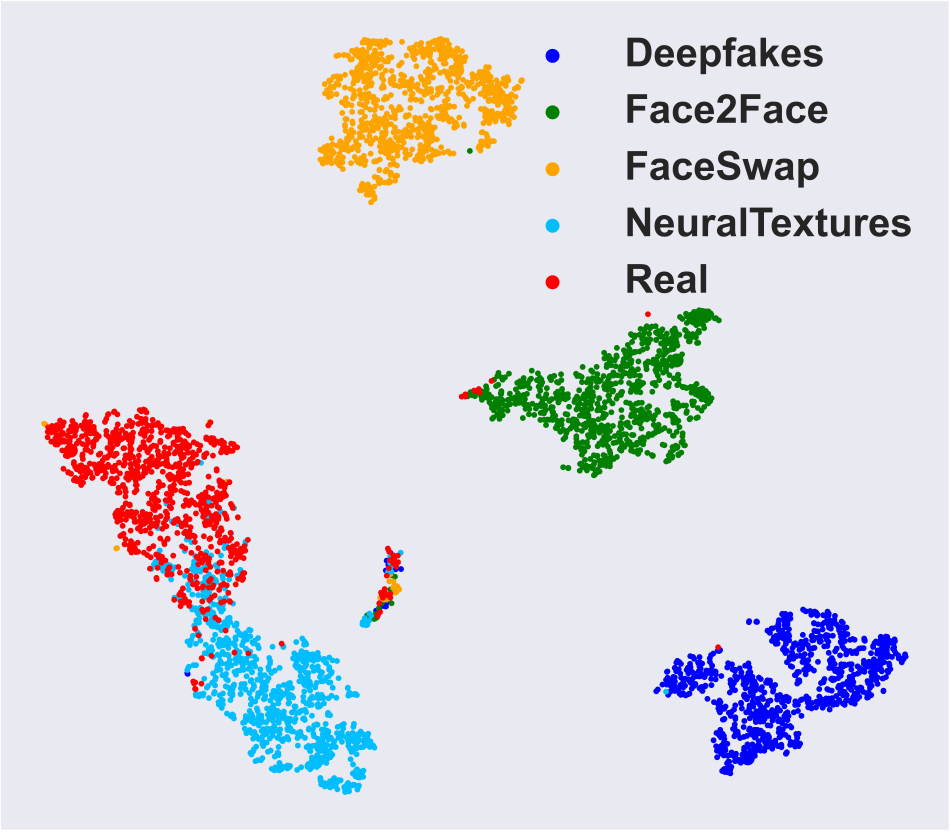}\label{tsne_dom}}
\caption{From left to right are visualizations of deepfake features of the EfficientNet-v2-L backbone network and our DID framework's deepfake features and domain features, respectively.
}
\label{fig:df_tsne}
\end{figure*}

\subsection{Visualization}
\noindent \textbf{Visualization of The Saliency Map.} 
To more intuitively demonstrate the effectiveness of our method, 
we visualize the Grad-CAM of deepfake attention $M_{df}$ and domain (forgery technique) attention $M_{dom}$ in Fig.~\ref{fig:attentionmap}. We can see from the figure that the activation regions of $M_{df}$ and $M_{dom}$ are different.
$M_{dom}$ focuses on areas such as the nose, mouth, and eyes.
In contrast, $M_{df}$ focuses on the information invariant to forgery techniques.
The visualization results demonstrate the effectiveness of our method: 
the decorrelation learning module promotes the disentangled components to
contain different information and be 
irrelevant to each other.

\noindent \textbf{Visualization of Deepfake and Domain Features.} 
Fig. \ref{backbone} and Fig. \ref{tsne_df}  depict the T-SNE visualization of the deepfake feature vectors 
learned by the backbone network EfficientNet-v2-L and our DID framework, respectively. 
The red and blue dots in the two figures represent the deepfake features of real and fake facial images, respectively. 
As illustrated in the figures, the real and fake features learned by the backbone network are mixed in the space,
whereas those learned by our DID framework are well separated in the feature space. 
This observation suggests the high real/fake discrimination performance of DID's deepfake features.

Fig.~\ref{tsne_dom} further presents a visualization of the domain features learned by our DID framework. 
We can see from the figure that the domain features learned from facial images created with different forgery techniques are well separated in the embedding space, with domain features from the same forgery technique being clustered together while those from other forgery techniques are far apart. 
These results demonstrate that the deepfake technique-related information is well
captured and separated in our DID framework.

\section{Conclusion}
We propose a deep information decomposition (DID) framework in this paper. 
It decomposes the deepfake facial information into deepfake-related and unrelated information, and further optimizes these two kinds of information to ensure they are sufficiently separated.
Only deepfake-related information is used for real/fake discrimination. 
 This approach makes the detection model robust to irrelevant changes and generalizable to unseen forgery methods.
Extensive experiments and visualizations demonstrate the effectiveness and superiority of the DID framework on cross-dataset deepfake detection tasks.

One limitation of our proposed DID framework is that all hyperparameters included in the loss function need to be manually selected through experiments. Another limitation is that our domain classification module needs to use domain information from the original deepfake dataset. However, obtaining such information from real-world datasets is often not easy.

For future work, our objective is to optimize these hyperparameters automatically during training. 
We also plan to design an auxiliary module to replace the current domain classification module without using the dataset's domain information.


\bibliographystyle{IEEEtran}
\bibliography{egbib}

\end{document}